\documentclass{preprint}

\usepackage{amsmath,amssymb,amsfonts}

\usepackage{algorithm}
\usepackage{algpseudocode}
\usepackage{amsmath}
\usepackage{booktabs}
\usepackage{tabularx}
\usepackage{makecell}
\usepackage{hyperref}
\usepackage[numbers,comma,sort&compress]{natbib}

\newcommand{\Input}[1]{\Statex \textbf{Input:} #1}
\newcommand{\Output}[1]{\Statex \textbf{Output:} #1}
\DeclareMathOperator*{\argmax}{arg\,max}

\usepackage{graphicx}
\usepackage{textcomp}
\usepackage{xcolor}
\usepackage[hypcap=false,width=\textwidth]{caption}

\begin{document}

\title{Suicide Risk Assessment on Social Media with Semi-Supervised Learning 
}

\author[1,2]{Max Lovitt}
\author[1]{Haotian Ma}
\author[3]{Song Wang}
\author[1]{Yifan Peng}

\affil[1]{Population Health Sciences, Weill Cornell Medicine, New York, New York, USA}
\affil[2]{The Masters School, Dobbs Ferry, New York, USA}
\affil[3]{Cockrell School of Engineering, The University of Texas at Austin, Austin, Texas, USA}
\affil[*]{Corresponding author(s). Email(s): \url{yip4002@med.cornell.edu}}

\maketitle

\begin{abstract}
With social media communities increasingly becoming places where suicidal individuals post and congregate, natural language processing presents an exciting avenue for the development of automated suicide risk assessment systems. However, past efforts suffer from a lack of labeled data and class imbalances within the available labeled data. To accommodate this task's imperfect data landscape, we propose a semi-supervised framework that leverages labeled (n=500) and unlabeled (n=1,500) data and expands upon the self-training algorithm with a novel pseudo-label acquisition process designed to handle imbalanced datasets. To further ensure pseudo-label quality, we manually verify a subset of the pseudo-labeled data that was not predicted unanimously across multiple trials of pseudo-label generation. We test various models to serve as the backbone for this framework, ultimately deciding that RoBERTa performs the best. Ultimately, by leveraging partially validated pseudo-labeled data in addition to ground-truth labeled data, we substantially improve our model’s ability to assess suicide risk from social media posts.
\end{abstract}


\section{Introduction}

Every year, more than 700,000 people worldwide die from suicide \cite{Organization2021-fi}. Tragically, this crisis is even more pronounced in adolescent populations: in 2021, suicide was determined to be the second leading cause of death for those aged 10 to 14 \cite{Pappas_undated-px}.

Suicide is a complex issue that poses no clear and consistent solution. The factors that drive individuals to commit suicide are complex and vary from individual to individual. The opioid crisis, economic recession, and access to firearms have all been named as potential contributors to this crisis~\cite{Marcotte2023-hl, Simon2024-vk}, but no single root cause can be isolated. Thus, targeting and supporting those at risk of suicide is crucial to combat this crisis, given that there is no simple solution to preventing suicidal ideation in the first place. 

Prior literature has consistently shown that suicide prevention interventions can reduce suicide attempts as well as completed suicides~\cite{Hofstra2020-xs}. Among these preventative methods, both pharmacological and social intervention methods have been established as effective~\cite{Lengvenyte2021-gm, Hou2022-ka}. Furthermore, restricting access to lethal means of suicide has also been shown to reduce suicide rates as well~\cite{Mann2005-nl}. 

With social media communities where individuals with suicidal thoughts congregate to seek advice increasing in size, researchers should scour these online communities as potential sites to deploy suicide intervention methods. However, due to massive amounts of traffic on the internet, manually inspecting every post, comment, and message for traces of suicidal ideation or intent is impossible. Thus, the key to establishing effective suicide prevention systems is automating suicide risk assessment. Such a system could play a crucial role in identifying individuals at risk of suicide and providing timely intervention and support. 

Prior efforts to automate this task have utilized deep learning and Natural Language Processing (NLP) to detect suicidal ideation in social media posts~\cite{Coppersmith2018-ei, Aldhyani2022-dz}. However, most previous works have constrained the problem to a binary classification task (i.e. “suicidal” versus “not suicidal”). This has resulted in the development of models that fail to capture varying levels of suicide risk (e.g. low risk, medium risk, high risk, etc.). Nonetheless, some progress has been made in constructing datasets that account for varied degrees of suicide risk. For example, some past efforts have involved fine-grained suicide risk datasets deriving different risk levels from the Columbia Suicide Severity Rating Scale~\cite{Gaur2019-ud, Li2022-tu}. However, these efforts have largely focused on constructing supervised models, learning strictly from the ground-truth labeled data. This means that the limited availability of labeled fine-grained suicide risk data incapacitates such models on the suicide risk assessment task~\cite{Ji2021-so}.

To bridge this gap, we investigated semi-supervised methods to learn from both labeled and unlabeled data. More specifically, we modified the existing self-training algorithm to optimize semi-supervised learning on highly-imbalanced datasets. After that, we compared the performances of supervised machine learning models to those utilizing our semi-supervised strategies. Through extensive empirical studies, we showed that the utilization of unlabeled data in a semi-supervised framework yielded substantial improvements in the task of identifying suicide risk levels in social media posts. 

To summarize, our contribution in this work is three-fold: (1) we proposed a novel semi-supervised learning framework with class-imbalanced data to generate pseudo-labeled data; (2) we conducted extensive empirical studies using the \textit{Suicide Detection on Social Media} BigData Cup Challenge \footnote{\url{https://www3.cs.stonybrook.edu/~ieeebigdata2024/}} dataset to compare an array of deep learning models’ ability in identifying the degrees of suicidal expressions in social media posts; (3) we manually validated a select subset of our pseudo-labeled data and evaluated its effect on improving model performance. 

\section{Materials and methods}
\label{sec:materials-methods}

\subsection{Task and Data Description}
\label{sec:task-data-description}

The \textit{Suicide Detection on Social Media} BigData Cup Challenge involves creating a predictive model to classify Reddit posts into four suicide risk levels: 1) Indicator: The post content has no explicit expression concerning suicide; 2) Ideation: The post content has explicit suicidal expression but there is no plan to commit suicide; 3) Behavior: The post content has explicit suicidal expression and a plan to commit suicide or self-harming behaviors; and 4) Attempt: The post content has explicit expressions concerning historic suicide attempts. 

As shown in Table \ref{tab:dataset}, the challenge organizers developed a corpus composed of 500 labeled and 1,500 unlabeled instances. A test dataset that includes 100 unlabeled instances is also included~\cite{Li2022-tu}. Each instance contains the text content of a Reddit post taken from the r/SuicideWatch sub-Reddit.

However, two major challenges are presented with this dataset. The first challenge is that only a small portion of available data is labeled, limiting the efficacy of a traditional supervised approach. The second challenge is that the intrinsic class-imbalance present in the labeled dataset makes it difficult for models trained on the provided data to predict classes underrepresented in the labeled corpus (i.e., Attempt).

\begin{table}
\caption{Statistics of the Suicide Detection on Social Media dataset.}
\label{tab:dataset}
\centering
\begin{tabular}{lr}
\toprule
Characteristics & $n$ (\%) \\
\midrule
Training & 500 \\
\hspace{1em}Indicator & 129 (25.8\%) \\
\hspace{1em}Ideation & 190 (38\%) \\
\hspace{1em}Behavior & 140 (28\%) \\
\hspace{1em}Attempt & 41 (8.2\%) \\
Testing & 100 \\
Unlabeled & 1,500 \\
\bottomrule
\end{tabular}%
\end{table}

\subsection{Method Development}
\label{sec:method-dev}

\begin{figure}
    \centering
    \includegraphics[width=\linewidth]{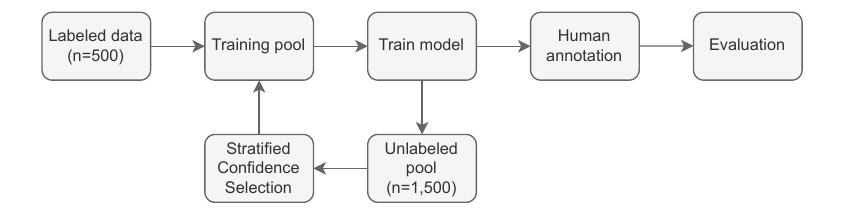}
    \caption{Proposed framework.}
    \label{fig:architecture}
\end{figure}

In this work, we hypothesized that a semi-supervised learning system utilizing the 1,500 unlabeled instances in addition to the 500 labeled posts could lead to better predictive performance than a supervised approach using only labeled data. An overview of the system architecture is shown in Figure \ref{fig:architecture}. The specific details of our backbone model and semi-supervised algorithm are described in the following subsections.

\subsubsection{Backbone Model Selection}
\label{sec:method-dev-supervised}

In this study, we employ a BERT-based classifier as the backbone within our semi-supervised framework. Specifically, we use RoBERTa~\cite{Liu2019-bh} to generate feature representations of input sentences given that previous literature has established that RoBERTa achieves strong performance on text classification tasks~\cite{Guo2020-nr}. We determined that RoBERTa outperformed other transformer-based models in this task, such as BERT base~\cite{Devlin2018-zv} and BiomedBERT~\cite{Gu2021-lz}.

\subsubsection{Semi-Supervised Learning with Stratified Confidence Sampling}
\label{sec:method-dev-scs}

Semi-supervised learning (SSL) is a family of machine learning algorithms designed to learn from a mix of both labeled and unlabeled data, typically in cases when the majority of the available data is unlabeled. Here, we employ the self-training strategy~\cite{Scudder1965-ts}, a semi-supervised wrapper algorithm wherein a classifier $f$ iteratively trains itself by generating a set of pseudo-labels $D_y$ for an unlabeled dataset $D_x$, adding a subset of those pseudo-labels $B_{y}$ -- typically selected by some confidence or certainty function $g$ -- to the original training set, and repeating this process until the model reaches optimal performance~\cite{Chapelle2010-rn}. 

\begin{algorithm}
\caption{Pseudocode for the Self-Training algorithm}
\label{alg:selftraining}
\begin{algorithmic}[1]
    \Input{${f(\bf{x})}$, $X_{i}$}, ${D}_{i}$, and ${g(\bf{y})}$
    \Output{${B}_{i}$ - Pseudo-labels}

    \State ${B}_{i} \gets [ ]$
    \While{${D}_{i} \neq \emptyset \land {b}_{i} \neq \emptyset$}
        \State Train ${f}$ on $X_{i} \cup {B}_{i}$
        \State $\bar{{Y_{i}}} \gets {f({D}_{i})}$

        \State ${b}_{i} \gets {g(\bar{{Y_{i}}})}$         
        
        \State ${B}_{i} \gets {B}_{i} \cup {b}_{i}$
        \State ${D}_{i} \gets {D}_{i} \setminus {b}_{i}$
    \EndWhile
    \State \textbf{return} ${B}_{i}$
\end{algorithmic}
\end{algorithm}

One crucial step to developing a task-specific self-training framework is to first establish what pseudo-labeled data should be added to the training pool with each iteration. One popular approach is to find the pseudo-labeled samples whose “confidence”, measured by predicted probabilities, exceeds a certain threshold~\cite{Yarowsky1995-gp}. However, this method struggles when dealing with imbalanced datasets. In our preliminary studies, we found that when using a confidence threshold to determine which samples to add to the training set almost all of the predictions exceeding that threshold would come from the class with the greatest number of instances in the initial labeled dataset. This problem would compound as more samples of the overrepresented class were iteratively added to the training set, resulting in an increasing confidence disparity between classes as well as diminishing performance. 

To combat this, we propose a Stratified Confidence Sampling (SCS) acquisition algorithm that utilizes model confidence as an indicator of pseudo-label accuracy while also ensuring that class imbalances in the initial training set do not hinder the model’s ability to select high-confidence instances of underrepresented classes. Specifically, SCS assigns pseudo-labels based on the prediction’s confidence relative to other samples with the same predicted label. This is done by separating the initial predictions by predicted class, ordering each subset based on predicted confidence, and then selecting the $p$ percent most-confident samples in each subset to be removed from the unlabeled set and added to the training set. The acquisition rate $p$ is a hyperparameter that, during the development of our model, we found worked best when set to 0.25. Note that $p$ should be within the range $(0, 1]$. 

\begin{algorithm}
\caption{Pseudocode for the Stratified Confidence Sampling algorithm}
\label{alg:scs}
\begin{algorithmic}[1]
    \Input{$X_{i}$ - posts; $Y_{i}^{n}$ - predicted probabilities; ${p}$ - acquisition rate}
    \Output{${B}_{x}$, ${B}_{y}$ - Set of posts with confident pseudo-labels}
    \State ${B}_{x} \gets [ ]$
    \State ${B}_{y} \gets [ ]$
    \For{For ${r}$ in range(${n}$)}
        \State ${y}_{i} \gets Y_{i}^{n}$ where $\argmax_i(Y_{i}^{n}, axis = {n}) = r$
        \State ${x}_{i} \gets X_{i}$ where $\argmax_i(Y_{i}^{n}, axis = {n}) = r$
        \State ${C_{i}} \gets \max_{i}\left(P(y_{i} = y_{1}), P(y_{i} = y_{2}),\dots, P(y_{i} = y_{n})\right)$
        \State ${N}_{r} \gets {p} * len({C_{i}})$
        \State ${y}_{i} \gets \text{Sort}({y}_{i} \text{ by } {C_{i}})$ 
        \State ${x}_{i} \gets \text{Sort}({x}_{i} \text{ by } {C_{i}})$
        \State ${B}_{x} \gets {B}_{x} + {x}_{i}[:{N}_{r}]$
        \State ${B}_{y} \gets {B}_{y} + {y}_{i}[:{N}_{r}]$
    \EndFor
    \State \textbf{return} {${B}_{x}, {B}_{y}$}
\end{algorithmic}
\end{algorithm}

\subsubsection{Label Correction}
\label{sec:method-dev-correction}

To ensure that the predicted pseudo-labels were as accurate as possible, we carried out our self-training algorithm five times, each time using a different subset of the ground-truth labeled data as a validation set. This process ultimately yielded five sets of pseudo-labels for the unlabeled dataset, and, to determine which label to use for each post, the majority vote was taken between the five sets of predictions. 

To further ensure pseudo-label accuracy, we manually verified the pseudo-label for all posts where the five initial pseudo-labels were not unanimous. Similar to Active Learning~\cite{Cohn1994-rq} in intuition, we use the models’ agreement to determine the samples that tend to “confuse” our classifier the most. We propose that labeling these samples is an efficient way to improve the quality of pseudo-labeled data while only manually validating a small portion of the pseudo-labeled dataset.

\subsection{LLMs with Zero-Shot Learning and Fine-Tuning}
\label{sec:method-llms}

During our preliminary experiments, we also tested the ability of LLMs (GPT-3.5-turbo and Llama3) to work as text classifiers within our semi-supervised framework. However, when attempting to feed the input posts into the GPT model, the model would repeatedly return an error, indicating that inappropriate content had been detected. As a result, we were not able to collect results using GPT-3.5-turbo and instead opted to use Llama3-8B.

We evaluated the capability of Llama3-8B to detect suicide risk levels in the zero-shot learning setting and fine-tuning setting. For fine-tuning, we compared two strategies: next-token prediction and sequence classification. For both zero-shot learning and fine-tuning, the prompt starts with the definitions of the four suicide risk levels, followed by the input text (see Supplementary Table~\ref{tab:prompt}).

\subsection{Evaluation Metrics}
\label{sec:eval-metrics}

During evaluation, we collected the micro and macro F1 scores to gauge the strength of the model’s predictions. Of these two metrics, micro-F1 is derived by taking the micro average of each label’s F1 score, which is computationally equivalent to accuracy, and macro-F1 is derived by averaging each class’ F1 score \cite{Grandini2020-cl}. Furthermore, we also collect precision, recall, and binary-F1 metric values for each class during the evaluation process to investigate model performance on a more granular level.

\subsection{Experimental Settings}\label{sec:exp settings}

When training BERT-based models, we used categorical cross entropy~\cite{Zhang2018-ak} as our loss function and trained each model using the Adam optimizer~\cite{Kingma2014-vh} with a batch size of 8 and a learning rate set to $10^{-5}$ for 10 epochs. Because most posts in both the labeled and unlabeled data sets consist of 200 or fewer tokens, we impose a maximum token length of 250 using the models' tokenizer’s built-in truncation and padding arguments to ensure uniform length between input sentences. Furthermore, we used the Inverse Class Frequency method to weight our loss function and handle the class imbalance issue. To reduce variability, we performed 5-fold cross-validation using different partitions of the data and reported precision, recall, micro and macro F1 scores. For evaluating model performance during each fold, we chose the model version with the highest accuracy on the validation set to represent our model's performance on that fold. To prevent our SCS algorithm from running for an absurd number of iterations -- given that the number of pseudo-labeled samples added to the initial set gets smaller and smaller with each iteration -- we chose to stop the SCS algorithm once the set of unlabeled posts reached below a certain threshold, which we set to 200. The remaining posts were then labeled using a classifier trained on both the labeled and pseudo-labeled sets. From here, we compiled all initially unlabeled posts into a dataset of 1,500 pseudo-labeled posts.

To fine-tune Llama3-8B, we loaded the model locally with 4-bit quantization~\cite{Dettmers2023-ae}, and, during the training process, we used Low-Rank-Adaptation (LoRA) with hyperparameters $a=8$ and $r=16$, training only a fraction of the model’s eight billion parameters~\cite{Hu2021-wc}. Furthermore, we used Adam with a learning rate of $2\times 10^{-4}$ -- which Dettmers et al.~\cite{Dettmers2024-po} indicates is optimal for quantized, low-rank adapted LLMs of this size -- and a cosine learning rate scheduler to optimize our model. 

\section{Results}

\subsection{RoBERTa Outperforms Other Models}
\label{sec:roberta}

\begin{figure}
    \centering
    \includegraphics[width=.5\linewidth]{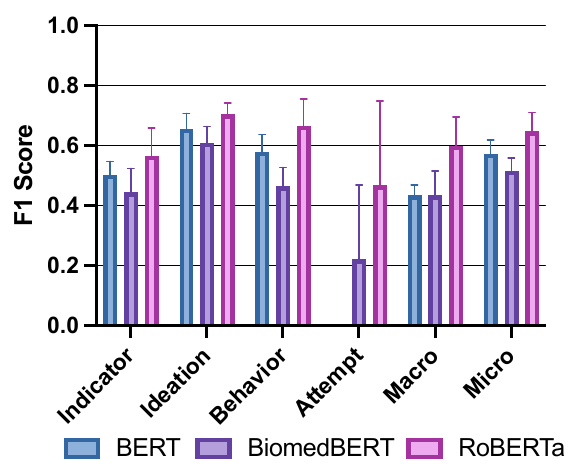}
    \caption{Comparing the F1 scores (both class-specific and micro/macro averaged) of three BERT-based models trained using the 500 labeled posts.}
    \label{fig:model selection}
\end{figure}

Before evaluating our semi-supervised approach, we first fine-tuned three transformer-based models using a supervised approach to determine which embedding model performs the best and establish a baseline against which we can compare our semi-supervised methods. Three transformer-based models, BERT~\cite{Devlin2018-zv}, BiomedBERT~\cite{Gu2021-lz}, and RoBERTa~\cite{Liu2019-bh} were trained and validated according to our experimental settings outlined in Section \ref{sec:exp settings}. Figure \ref{fig:model selection} shows the five-fold cross-validated performance of our systems. In columns labeled "Indicator", "Ideation", "Behavior", and "Attempt", we plot the binary-F1 score corresponding to posts of that class. The F1 scores yielded by our supervised models indicate that RoBERTa outperformed the other two embedding models across all F1 metrics (both class-specific and multi-class). This improvement is especially pronounced when examining the class-specific metrics. A more detailed report comparing the performance between the three models can be found in Supplementary Table \ref{tab:model selection}.

During the evaluation, RoBERTa achieved an average micro F1 score of 0.648 (standard deviation: 0.061) and an average macro F1 score of 0.599 (standard deviation: 0.970). This represents a substantial improvement in comparison to BERT, which achieves average micro and macro F1 scores of 0.572 (standard deviation: 0.046) and 0.434 (standard deviation: 0.034). Furthermore, RoBERTa also outperformed BiomedBERT, which obtained an average micro F1 score of 0.516 (standard deviation: 0.043) and an average macro F1 score of 0.434 (standard deviation: 0.080). 

RoBERTa also exceeded the other two models in precision, recall, and binary-F1 scores, with the most noticeable improvement occurring in the F1 score for the “Attempt” class. For samples labeled as “Attempt”, RoBERTa achieves an F1 score of 0.467 (standard deviation: 0.280) in comparison to BERT’s 0.000 and BiomedBERT’s 0.212 (standard deviation: 0.250). Aside from outperforming the other two models across the board, another reason behind selecting RoBERTa over BERT and BiomedBERT is the latter two models’ abysmal performance in classifying posts within the “Attempt” class. As a result, for the remainder of our experiments, we used RoBERTa as the backbone for our semi-supervised approaches.

After deciding on RoBERTa as an embedding model, we conducted self-training with SCS pseudo-label acquisition to generate a dataset of 1,500 pseudo-labeled posts according to the methods outlined in Section \ref{sec:method-dev-scs}. Furthermore, we also partially verified these pseudo-labels according to the strategy outlined in Section \ref{sec:method-dev-correction}.

\subsection{Correcting Initial Pseudo-Labels}\label{sec:results-correcting}

\begin{figure}
    \centering
    \includegraphics[width=0.35\linewidth]{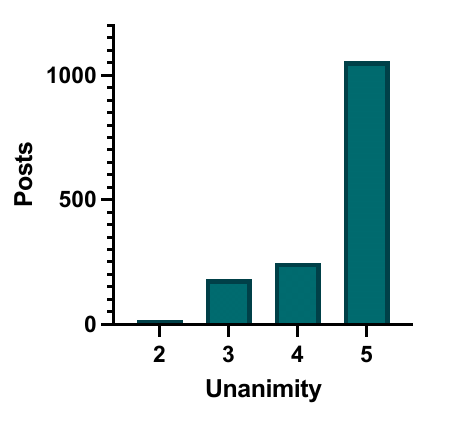}
    \vspace{-1em}
    \caption{Distribution of unanimous and non-unanimous pseudo-labels. The unanimity values (2, 3, 4, and 5) along the x-axis represent how many votes the majority-vote pseudo-label received for that instance.}
    \label{fig:count}
\end{figure}

\begin{figure}
    \centering
    \includegraphics[width=.5\linewidth]{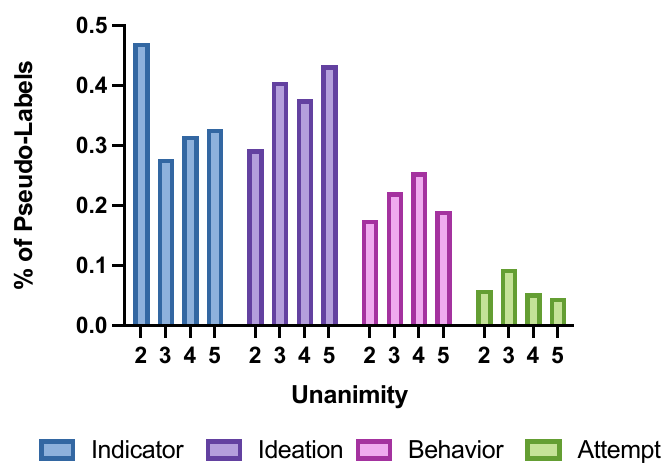}
    \caption{Relative class frequencies for each level of pseudo-label unanimity.}
    \label{fig:freq}
\end{figure}

After training five models and generating five sets of pseudo-labels, we calculated the agreement between each set of pseudo-labels and the majority-vote pseudo-labels using Cohen’s Kappa score \cite{Cohen1960-tg} (Supplementary Table \ref{tab:kappa}). The rate of agreement between the five sets of pseudo-labels is high, which makes sense given that the training data between any two of the five models is 75\% similar. To determine which samples to annotate, the final set of 1,500 pseudo-labeled posts was first split into four separate subsets based on the number of votes the majority voted class received between the 5 initial pseudo-labelings. Figure \ref{fig:count} details the number of samples for each level of unanimity, and Figure \ref{fig:freq} relative class frequencies for each level of unanimity. The numerical data displayed in this figure can be found in Supplementary Table \ref{tab:unanimity}.

Two annotators verified the 444 non-unanimous pseudo-labels, labeling each pseudo-label as either correct or incorrect. Of the 444 non-unanimous pseudo-labels, the first 104 were verified by both validators to ensure agreement on how to validate labels. Within these 104 posts, the validators agreed in 95.2\% of cases. From there, the remaining 340 posts were then split into groups of 170, which were validated independently. 

Between the two validators, 287 of the 444 posts were determined to be correctly labeled (64.6\% pseudo-label accuracy). Furthermore, all samples marked as incorrectly labeled were then assigned new labels according to their correct classification. Both the uncorrected pseudo-labeled dataset as well as the partially corrected pseudo-labeled dataset were used as additional training data to supplement the initial 500 labeled posts to determine how the addition of this pseudo-labeled data affects model performance.

\subsection{Training with Pseudo-Labeled Data}\label{sec:results-pl-data}

Each pseudo-labeled dataset (n=1,500) was used as additional training data, and we evaluated RoBERTa trained on the labeled and pseudo-labeled data (n=2,000) using five-fold-cross-validation. Note that in our validation scheme, we chose to only sample validation data from the initial set of ground-truth labeled posts, not the pseudo-labeled data. This was done to prevent the model from verifying its performance against potentially incorrect labels.

\begin{figure}
    \centering
    \includegraphics[width=.5\linewidth]{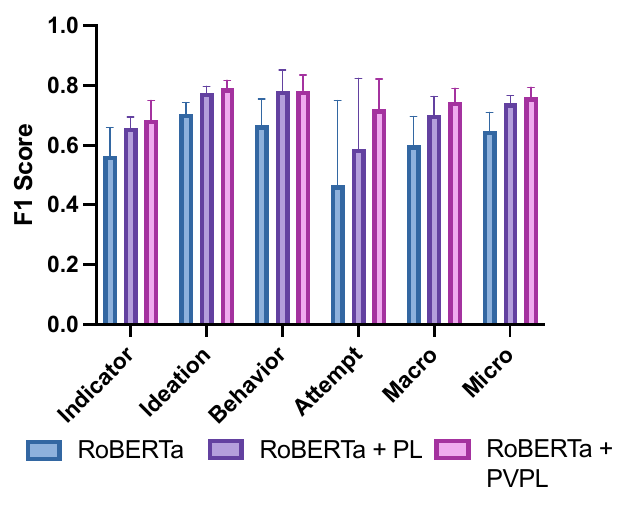}
    \caption{F1-scores for RoBERTa trained with pseudo-labeled data (PL) and partially-validated pseudo-labeled data (PVPL).}
    \label{fig:experimental}
\end{figure}

The cross-validated F1 metrics for utilizing the two sets of pseudo-labeled data according to this strategy are shown in Figure \ref{fig:experimental}. We also chose to include the F1 metrics of RoBERTa trained purely on the 500 labeled posts to serve as a point of comparison. A more detailed report comparing the three frameworks can be found in Supplementary Table \ref{tab:roberta}.

As indicated by the data, the addition of both the unverified and partially-verified pseudo-labels to the model’s training data archives impressive boosts in performance, resulting in an average micro-F1 score of 0.760 (standard deviation: 0.034) and an average macro-F1 score of 0.745 (standard deviation: 0.046) for RoBERTa trained on the 500 labeled posts as well as the 1,500 partially validated pseudo-labeled data. Furthermore, this improvement is also noticeable when looking at class-specific metrics, with this effect being the most pronounced for the class least represented in the initially labeled dataset, “Attempt”. The addition of the unverified and partially verified pseudo-labeled data boosts the F1-Attempt score from 46.7\% (standard deviation: 0.282) to 58.9\% (standard deviation: 0.235) and 72.0\% (standard deviation: 0.102) respectively. This constitutes an improvement in F1-score by 25.3\% for RoBERTa trained on both the labeled and partially verified pseudo-labeled data.

After submitting our model weights to the organizers of the IEEE \textit{Sucide Detection on Social Media} Big Data Cup Challenge, they evaluated our model on a 100-post test set, using weighted F1 as their primary metric. We ranked fourth out of 21 participating teams in terms of weighted F1 score (0.7463). In addition to purely evaluation metrics, the competition organizers also compiled a final score for each team, based on approach innovation, written report quality, and final weighted F1 score. When judged by this metric, we ultimately ranked second out of the participating 21 teams.

\subsection{Lessons Learned from LLMs}
\label{sec:llms}

In addition to evaluating the performance of transformer-based models, we also chose to evaluate Llama3-8B as described in Section \ref{sec:method-llms}. Given that Llama3-8B is very time-consuming to fine-tune and that our results indicated Llama3-8B performed worse than RoBERTa, we did not conduct five-fold cross-validation. Therefore the results shown in Figure \ref{fig:llm} reflect only one training fold.  A more detailed classification report can be found in Supplementary Table \ref{tab:Llama3}.

\begin{figure}
    \centering
    \includegraphics[width=0.5\linewidth]{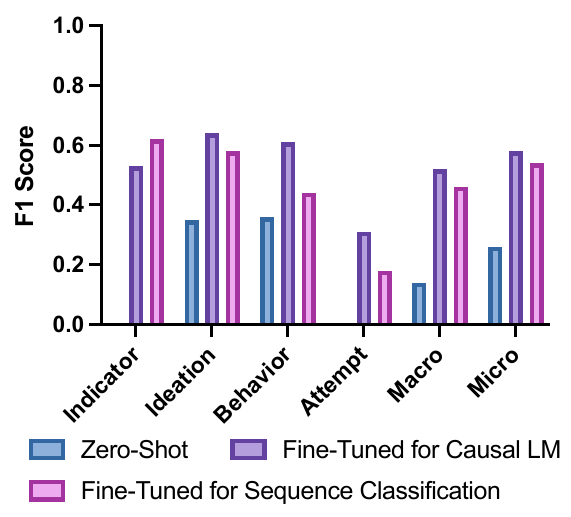}
    \caption{Comparing F1 scores for zero-shot and fine-tuned vanilla Llama3-8B.}
    \label{fig:llm}
\end{figure}

Our results indicate that Llama3’s performance under zero-shot circumstances is similar to random chance; however, for both fine-tuning strategies, Llama3 shows an ability to assess suicide risk competitive with BERT and BiomedBERT, with fine-tuning for Causal LM achieving better performance than Sequence Classification. 

\section{Discussion}
\label{sec:disc}

Our results indicate that the addition of pseudo-labeled data can substantially improve model performance for deep learning architectures. This insight is increasingly relevant as emerging deep-learning models require ever more data to train; however, supplementing human-annotated ground-truth labels with model-generated pseudo-labels serves as a promising path forward to alleviate this issue. 

The results from our experiment, displayed in Section \ref{sec:results-pl-data}, indicate that self-training with SCS pseudo-label acquisition improves model performance by a substantial margin. Furthermore, the additional step of validating the posts that are not unanimously pseudo-labeled across an ensemble of models provides an additional boost to classification accuracy; however, the majority of the improvement in model performance comes from the addition of the pseudo-labeled data, not the partial-validation. While our proposed SCS strategy exhibits empirical success, it is important to note that our method relies on the assumption that the class distribution in the unlabeled data is equivalent or near-equivalent to that of the labeled data, since the class distribution of the model’s predictions, and thus the distribution of our pseudo-labeled data, is closely tied to the class distribution in our initial dataset. Future deployments of this method should be mindful of this assumption.

Additionally, the data we record in Section \ref{sec:results-pl-data} suggests that self-training with SCS pseudo-label acquisition disproportionately boosts F1 scores in classes underrepresented in the initial dataset. This indicates that, in addition to addressing the shortcomings presented by traditional pseudo-label selection techniques when dealing with imbalanced datasets, our SCS strategy is also able to partially correct the disparity in model performance between imbalanced classes. Therefore, our SCS strategy also presents a promising method for addressing class-imbalanced partially labeled tasks beyond just suicide risk assessment that we urge future researchers to validate and explore. This effect is even more pronounced for RoBERTa trained on both the labeled and partially validated pseudo-labeled data. Section \ref{sec:results-correcting} details the class distribution of the unanimous versus the non-unanimous pseudo-labels and demonstrates that the set of “confusing” samples contains a greater percentage of the “Attempt” class than the unanimously labeled posts. Given the large improvement in the F1 score for the “Attempt” class for RoBERTa trained on the corrected pseudo-labels, we believe correcting the pseudo-labels that tend to confuse the pseudo-labeling algorithm presents a way to improve model performance in predicting underrepresented classes, as those underrepresented classes tend to be the ones confusing the model most frequently.

Furthermore, our results in Section \ref{sec:llms} compare the performance of Llama3-8B to RoBERTa. While we ultimately found that RoBERTa achieved superior performance and was computationally cheaper, comparing the performance of LLMs to BERT-based models is an important effort to gauge the ability of these models to perform non-generative tasks. Our finding that RoBERTa outperforms Llama3-8B on text-classification tasks is consistent with prior literature \cite{Yu2023-io} and poses an interesting question about the strengths and weaknesses of LLMs. However, given the specific nature of this task, our findings should not be treated as an absolute indicator of Llama3’s text-classification abilities, and we implore future researchers to investigate the viability of LLMs in other text-classification tasks.

Although our SCS strategy yields substantial improvements over a baseline RoBERTa classifier, there is a clear margin for improvement. The simplest improvement to our research would be to increase the size of the unlabeled dataset. Given that a pseudo-labeled dataset of size 1500 was able to increase model accuracy by over 10 percent, leveraging an additional 500, 1000, or even 1500 more unlabeled posts could further improve model performance. Future research should determine the optimal amount of unlabeled data to use by investigating at what point, if at all, the addition of pseudo-labeled data results in diminishing returns in model performance. 

Furthermore, comparing our SCS strategy to other semi-supervised strategies, especially for training on class-imbalanced datasets, would serve as a helpful tool for future researchers dealing with large amounts of unlabeled data to choose the best method for their task.

\section{Conclusions}

Faced with the growing suicide crisis, we proposed and implemented a new self-training strategy to develop automated methods of suicide risk assessment for social media posts when the available labeled data is both limited and imbalanced. We evaluated the strategy on the dataset made available through the IEEE BigData 2024 Cup task, \emph{Suicide Detection on Social Media}. The results of our evaluation demonstrate that self-training with SCS pseudo-label acquisition can effectively detect varying suicide risk levels from Reddit posts with cutting-edge performance. In addition, we investigated the effect of human annotation on a subset of our pseudo-labeled and demonstrated that it can provide even further improvements over baseline models. While our work only scratches the surface of semi-supervised learning in the suicide domain, we hope it will shed light on the development of new frameworks that can lead to gains from human-in-the-loop.

\paragraph{Code Availability}

The notebooks we used to pseudo-label our data and evaluate our models are stored in the GitHub repository linked here. \url{https://github.com/bionlplab/2024_ieee_big_data_suicide_ideation_detection/tree/main}

\paragraph{Acknowledgments}

This work was supported by the AIM-AHEAD Consortium Development Program of NIH under grant number OT2OD032581 and the National Science Foundation under grant number 2145640.

\paragraph{Author Contributions}

M.L, Y.P contributed to the conception of the study and study design; Y.P. contributed to the acquisition of the data; M.L. contributed to the analysis and interpretation of the data; M.L., H.M. contributed to the data annotation; all contributed to drafting and revising the manuscript.

\paragraph{Competing Interests}

The authors declare no competing interests.

\bibliographystyle{IEEEtran}
\bibliography{ref}

\clearpage

\newpage
\onecolumn
\renewcommand*{\appendixname}{Supplementary materials}


\appendix
\setcounter{table}{0}
\setcounter{figure}{0}
\renewcommand\figurename{Supplementary Figure} 
\renewcommand\tablename{Supplementary Table}

\begin{center}
\captionof{table}{Prompt for Llama3}
\label{tab:prompt}
\centering
\begin{tabularx}{.8\linewidth}{X}
\toprule
Here are four different post classification options: 0: Indicator, the post content has no explicit expression concerning suicide, 1: Suicidal ideation, the post content demonstrates explicit suicidal expression but there is no plan to commit suicide, 2: Suicidal behavior, the post content has explicit suicidal expression and a plan to commit suicide or self-harming behaviors, 3: Suicide attempt, the post content has explicit expressions concerning historic suicide attempts. Which of those labels best describes the following text: \textit{Post content}. Answer: \\
\bottomrule
\end{tabularx}
\end{center}

\newpage

\begin{center}
\captionof{table}{Detailed classification report for BERT, BiomedBERT, and RoBERTa trained on 500 labeled instances.}
\label{tab:model selection}
\resizebox{\textwidth}{!}{%
\begin{tabular}{l *9{r@{~(}r@{)~~~}}}
\toprule
& \multicolumn{6}{c}{BERT} & \multicolumn{6}{c}{BiomedBERT} & \multicolumn{6}{c}{RoBERTa} \\
\cmidrule(rl){2-7} \cmidrule(rl){8-13} \cmidrule(rl){14-19}
 & 
  \multicolumn{2}{r}{Precision~~~} 
 & \multicolumn{2}{r}{Recall~~~} 
 & \multicolumn{2}{r}{F1~~~} 
 & \multicolumn{2}{r}{Precision~~~} 
 & \multicolumn{2}{r}{Recall~~~} 
 & \multicolumn{2}{r}{F1~~~} 
 & \multicolumn{2}{r}{Precision~~~} 
 & \multicolumn{2}{r}{Recall~~~} 
 & \multicolumn{2}{r}{F1~~~} \\
\midrule
Indicator & 0.472 & 0.096 & 0.607 & 0.208 & 0.502 & 0.045 & 0.402 & 0.111 & 0.540 & 0.166 & 0.444 & 0.079 & 0.496 & 0.099 & 0.664 & 0.083 & 0.566 & 0.093 \\
Ideation & 0.774 & 0.121 & 0.576 & 0.045 & 0.655 & 0.052 & 0.726 & 0.155 & 0.539 & 0.048 & 0.609 & 0.056 & 0.753 & 0.076 & 0.666 & 0.026 & 0.705 & 0.038 \\
Behavior & 0.557 & 0.092 & 0.612 & 0.046 & 0.579 & 0.058 & 0.429 & 0.088 & 0.526 & 0.118 & 0.464 & 0.063 & 0.700 & 0.086 & 0.646 & 0.129 & 0.666 & 0.088 \\
Attempt & 0.000 & 0.000 & 0.000 & 0.000 & 0.000 & 0.000 & 0.200 & 0.227 & 0.248 & 0.273 & 0.221 & 0.247 & 0.467 & 0.303 & 0.467 & 0.283 & 0.467 & 0.282 \\
micro & - & - & - & - & 0.572 & 0.046 & - & - & - & - & 0.516 & 0.043 & - & - & - & - & 0.648 & 0.061 \\
macro & - & - & - & - & 0.434 & 0.034 & - & - & - & - & 0.434 & 0.080 & - & - & - & - & 0.599 & 0.097\\
\bottomrule
\end{tabular}}
\end{center}

\newpage

\begin{center}
\captionof{table}{Agreement between 5 models and majority vote pseudo-labels}
\label{tab:kappa}
\centering
\begin{tabular}{cr}
\toprule
Model & Cohen’s Kappa\\
\midrule
1 & 0.878 \\
2 & 0.864 \\
3 & 0.878 \\
4 & 0.895 \\
5 & 0.837 \\
\bottomrule
\end{tabular}
\end{center}

\newpage

\begin{center}
\captionof{table}{Distribution in unanimity and class between five sets of pseudo-labels}
\label{tab:unanimity}
\centering
\begin{tabular}{l *{4}{r@{~~}r@{\%~~~}}}
\toprule
 & \multicolumn{2}{c}{2} & \multicolumn{2}{c}{3} & \multicolumn{2}{c}{4} & \multicolumn{2}{c}{5} \\
 \midrule
Number of posts & 17 & \multicolumn{1}{c}{-} & 180 & \multicolumn{1}{c}{-} & 247 & \multicolumn{1}{c}{-} & 1056 & \multicolumn{1}{c}{-} \\
Indicator & 8 & 47.1 & 50 & 27.8 & 78 & 31.6 & 347 & 32.8 \\
Ideation & 5 & 29.4 & 73 & 40.6 & 93 & 37.7 & 458 & 43.4 \\
Behavior & 3 & 17.6 & 40 & 22.2 & 63 & 25.5 & 202 & 19.1 \\
Attempt & 1 & 5.9 & 17 & 9.4 & 13 & 5.3 & 49 & 4.6\\
\bottomrule
\end{tabular}

\newpage

\captionof{table}{Detailed classification report for RoBERTa trained on labeled data, labeled and pseudo-labeled (PL) data, and labeled and partially validated pseudo-labeled (PVPL) data.}
\label{tab:roberta}
\resizebox{\textwidth}{!}{%
\begin{tabular}{l *9{r@{~(}r@{)~~~}}}
\toprule
 & \multicolumn{6}{c}{RoBERTa Fine-tuned} & \multicolumn{6}{c}{RoBERTa Fine-tuned + Pseudo Labels} & \multicolumn{6}{c}{\makecell[c]{RoBERTa Fine-tuned \\+ Partially-verified Pseudo Labels}} \\
 \cmidrule(rl){2-7} \cmidrule(rl){8-13} \cmidrule(rl){14-19}
 & 
 \multicolumn{2}{r}{Precision~~~} 
 & \multicolumn{2}{r}{Recall~~~} 
 & \multicolumn{2}{r}{F1~~~} 
 & \multicolumn{2}{r}{Precision~~~} 
 & \multicolumn{2}{r}{Recall~~~} 
 & \multicolumn{2}{r}{F1~~~} 
 & \multicolumn{2}{r}{Precision~~~} 
 & \multicolumn{2}{r}{Recall~~~} 
 & \multicolumn{2}{r}{F1~~~} \\
 \midrule
Indicator & 0.496 & 0.099 & 0.664 & 0.083 & 0.566 & 0.093 & 0.792 & 0.075 & 0.574 & 0.076 & 0.658 & 0.035 & 0.814 & 0.057 & 0.605 & 0.107 & 0.685 & 0.065 \\
Ideation & 0.753 & 0.076 & 0.666 & 0.026 & 0.705 & 0.038 & 0.695 & 0.035 & 0.879 & 0.027 & 0.775 & 0.022 & 0.729 & 0.038 & 0.868 & 0.041 & 0.792 & 0.026 \\
Behavior & 0.700 & 0.086 & 0.646 & 0.129 & 0.666 & 0.088 & 0.831 & 0.065 & 0.750 & 0.113 & 0.783 & 0.069 & 0.790 & 0.041 & 0.779 & 0.069 & 0.783 & 0.052 \\
Attempt & 0.467 & 0.303 & 0.467 & 0.283 & 0.467 & 0.282 & 0.690 & 0.291 & 0.581 & 0.284 & 0.589 & 0.235 & 0.791 & 0.206 & 0.683 & 0.057 & 0.720 & 0.102 \\
micro & - & - & - & - & 0.648 & 0.061 & - & - & - & - & 0.740 & 0.025 & - & - & - & - & 0.760 & 0.034 \\
macro & - & - & - & - & 0.599 & 0.097 & - & - & - & - & 0.701 & 0.062 & - & - & - & - & 0.745 & 0.046\\
\bottomrule
\end{tabular}
}
\end{center}

\newpage

\begin{center}
\captionof{table}{Detailed classification report for Llama3 8-B zero-shot evaluated on the 500 labeled instances as well as Llama3 8-B fine-tuned on the 500 labeled instances for Causal LM and Sequence Classification.}
\label{tab:Llama3}
\centering
\begin{tabular}{lrrrrrrrrr}
\toprule
 & \multicolumn{3}{c}{Zero Shot} & \multicolumn{3}{c}{\makecell[c]{Fine Tuned for \\Causal LM}} & \multicolumn{3}{c}{\makecell[c]{Fine Tuned for \\Sequence Classification}} \\
 \cmidrule(rl){2-4} \cmidrule(rl){5-7} \cmidrule(rl){8-10}
 & Precision & Recall & F1 & Precision & Recall & F1 & Precision & Recall & F1 \\
\midrule
Indicator & 0.00 & 0.00 & 0.00 & 0.63 & 0.46 & 0.53 & 0.55 & 0.71 & 0.62 \\
Ideation & 0.46 & 0.29 & 0.35 & 0.56 & 0.74 & 0.64 & 0.55 & 0.61 & 0.58 \\
Behavior & 0.27 & 0.54 & 0.36 & 0.61 & 0.61 & 0.61 & 0.53 & 0.38 & 0.44 \\
Attempt & 0.00 & 0.00 & 0.00 & 0.50 & 0.22 & 0.31 & 0.50 & 0.11 & 0.18 \\
micro & - & - & 0.26 & - & - & 0.58 & - & - & 0.54 \\
macro & - & - & 0.14 & - & - & 0.52 & - & - & 0.46\\
\bottomrule
\end{tabular}
\end{center}

\end{document}